# Optimal Blackjack Strategy Recommender:
# A Comprehensive Study on Computer Vision Integration for Enhanced Gameplay

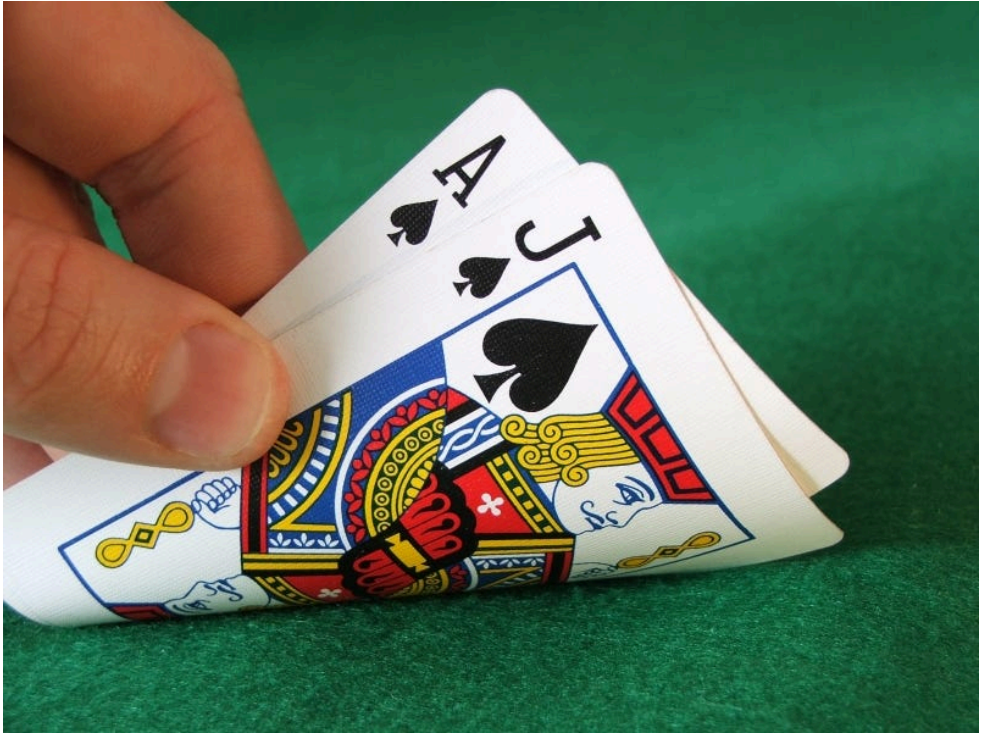

Cal Poly
EE 428 - Research Project
March 23, 2024

Krishnanshu Gupta
Devon Bolt
Ben Hinchliff

# 1 Abstract

This research project investigates the application of several computer vision techniques for playing card detection and recognition in the context of the popular casino game, blackjack. The primary objective is to develop a robust system that is capable of detecting and accurately classifying playing cards in real-time, and displaying the optimal move recommendation based on the given image of the current game. The proposed methodology involves using K-Means for image segmentation, card reprojection and feature extraction, training of the KNN classifier using a labeled dataset, and integration of the detection system into a Blackjack Basic Strategy recommendation algorithm. Further, the study aims to observe the effectiveness of this approach in detecting various card designs under different lighting conditions and occlusions. Overall, the project examines the potential benefits of incorporating computer vision techniques, with a specific focus on card detection, into commonly played games aiming to enhance player decision-making and optimize strategic outcomes. The results obtained from our experimental evaluations with models developed under considerable time constraints, highlight the potential for practical implementation in real-world casino environments and across other similarly structured games.

# 2 Introduction and Data Description

Blackjack stands as one of the most popular and strategic casino card games, with the lowest house edge at 1.5-2% depending on the specific configuration of the game. Additionally, there is a fully computed optimal strategy for Blackjack, called 'Basic Strategy' which is derived from extensive mathematical and statistical analysis. However, employing and memorizing this Basic Strategy can be challenging for novice players, as it requires quick and precise interpretation of the player and dealer cards. Motivated by a desire to empower and train novice players and help them bridge the skill gap to gain a competitive edge against the casino and other players, we explored the integration of computer vision techniques into Blackjack gameplay.

The problem statement and objective of this project is to take an image of a blackjack game, accurately and efficiently detect the playing cards, classify the images and determine them as player or dealer cards, and recommend the optimal move for the player. This involved overcoming several obstacles and implementing different techniques such as reprojection and feature extraction for detecting the cards, K-Means for image segmentation, and KNN classifier to identify the specific cards. Through these approaches, we aimed to share and democratize access to strategic gameplay insights to all players, thereby enhancing the Blackjack experience for players within the novice level.

To facilitate and achieve this goal, we first created a comprehensive dataset consisting of a diverse range of single playing card images, multi-playing card images, and various blackjack game setups. Furthermore, we included more challenging images tailored for testing and evaluation purposes of the card detection and recognition models. These images were all taken on one of our cell phones, as this allowed for the images to be consistently high-resolution with a minimum of compression artifacts and limits the size of data collection due to time constraints. The single playing card images were individually labeled with ground truth annotations, indicating each card's identity, which were necessary for training the KNN classifier. Initially, we explored the possibility of utilizing images from past research projects via platforms like Kaggle and online poker games due to their convenience. However, we encountered significant limitations with this as the available datasets often had undesirable compression artifacts or lacked the requisite high-resolution. Additionally, some images obtained online failed to meet our predefined assumptions, and as a result, we opted for our custom cell phone-captured images to ensure data quality.

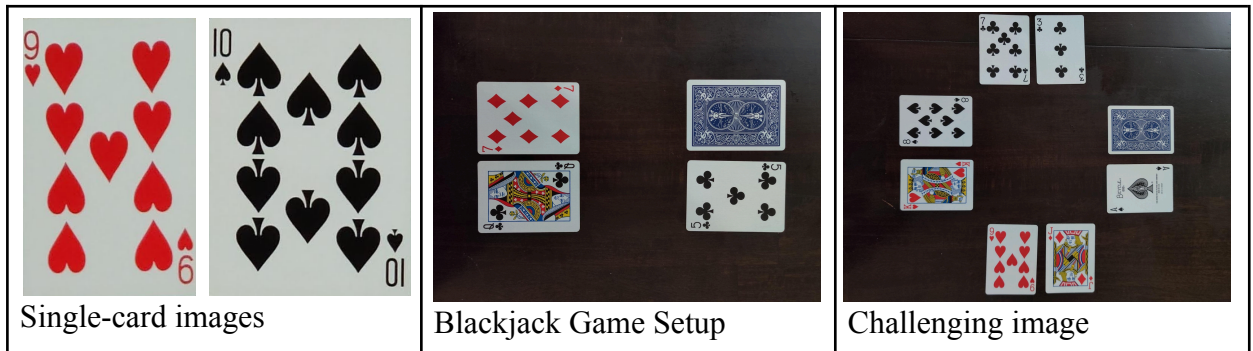

*Figure 1:* *Examples of Image Types in Dataset*

Considering the complexity and the time constraints of the project, we operated under several assumptions to streamline our approach and speed up model development and training timelines. First, we assumed that the cards would be from a standard pack of Bicycle playing cards, ensuring uniformity in design and size. Additionally, we assumed that all the cards would be in full view with no overlapping between them, which is done to make detection of distinct connected-components or regions of white pixels within the image possible. Furthermore, we relied on good lighting conditions with minimal glare to ensure clear visibility of the cards, and for a contrasting background so the cards could be easily separated from the background. The latter assumption is reasonable as the majority of blackjack tables at casinos have dark-colored backgrounds and designs. By adhering to these assumptions, we were able to optimize our model development, and successfully meet our defined project objectives in the given timeframe.

# 3 Related Work from Literature

Using computer vision to detect playing card values is not a new idea. Many people have implemented their own version of this algorithm and used it for various card games. However, every implementation we could find used deep learning based computer vision. While this technique has its benefits, mainly that it is easy to implement, it runs quickly, and it has a very high accuracy even for difficult data, once the models are trained, it doesn't allow us to understand what is going on and how the algorithm actually works.

A Medium[1] article published in 2023 details the process for using OpenCV and a library called YOLO (You Only Look Once) to implement a convolutional neural network based playing card detector for poker. This article focuses less on the technical details of the algorithms and more on the general concepts and how it can be applied to the game of poker. Overall, this source wasn't very useful to us.

A more formal-looking project was from an IEEE paper from 3 students at Stanford[2]. This paper details both classical and deep learning techniques and compares them. The two classical techniques discussed are Multiclass Logistic Regression and Multiclass Support Vector Machine. While it was interesting to read about these techniques, they were beyond the scope of this class so we decided not to use them and instead focus on the techniques we have learned about. Interestingly, this source points out that even though they used sophisticated classical computer vision techniques, the final results weren't nearly as good as with the deep learning based approaches.

Another interesting yet somewhat useful article was found on blog.collectors.com. In their article, "Automating Card Identification Using Computer Vision,"[3] the author discussed how he wanted to automatically classify collectible baseball cards using computer vision. He detailed how he first tried using classical computer vision techniques like SIFT, SURF, and ORB but kept falling short of the results he needed. Instead, he turned to convolutional neural networks to solve his problem. Since collectible cards are much more complicated than playing cards, we were hopeful that we could use classical techniques to detect playing cards. This also helped us realize that we would have to do a lot of preprocessing to the images in order for the feature detection and classification algorithms to work well.

---

[1] Reference [1]
[2] Reference [2]
[3] Reference [3]

# 4    Implementation

The core algorithm utilized in our project is referred to by us as CARP (Card Analysis and Recognition Program). The algorithm has 3 main parts: card detection, reprojection, and classification at which point the output can be fed into the rules-based blackjack algorithm. This algorithm was implemented using Python and OpenCV[4], a large open-source computer vision framework that provides a lot of pre-existing tools to ease development, similar to MATLAB.

The first step of card detection is to segment the cards from the background. Initially, this was done with global thresholding techniques, but these struggled to properly segment the background from the foreground due to variation in lighting and background. Since the cards are most easily identified as clusters of white pixels, we decided to use K-means clustering to isolate them, selecting the best cluster as the cluster with the highest luminance and the lowest saturation. K=3 clusters was chosen because it works quite well at creating one cluster for the background, one for the white sections of the cards, and one for the red text on some cards. Figure 2 shows an example of the image after K-means segmentation. Code 1 shows the code used for the segmentation part of the algorithm.

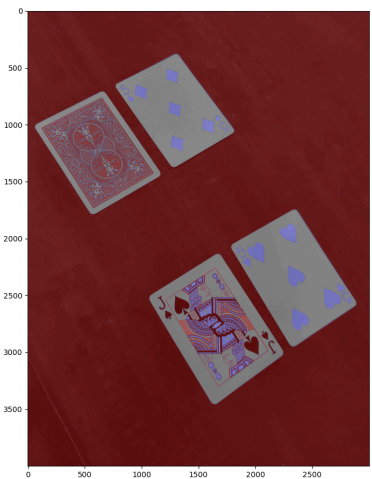

*Figure 2:* K-means clusters overlaid on original image of blackjack hand

```python
def segment_cards(img, K = 3):
    # Define criteria = ( type, max_iter = 10 , epsilon = 1.0 )
    criteria = (cv.TERM_CRITERIA_EPS + cv.TERM_CRITERIA_MAX_ITER, 10, 1.0)

    # Set flags (Just to avoid line break in the code)
    flags = cv.KMEANS_PP_CENTERS
```

---

[4] Reference [4]

```python
# Apply KMeans
w, h, c = img.shape
img_flat = np.float32(img.reshape(w * h, c))
_, labels, _ = cv.kmeans(img_flat, K, None, criteria, 10, flags)

# Convert image to HSV
img_hsv = cv.cvtColor(img, cv.COLOR_RGB2HSV).reshape((w * h, 3))
scores = []
for i in range(K):
labels = labels.reshape(-1)
mask = labels == i
masked = img_hsv[mask]
# Calculate "score" for each cluster as (1.0 - average saturation) + average luminance
avg_sat = masked[:, 1].mean()
avg_val = masked[:, 2].mean()
scores.append((255.0 - avg_sat) + avg_val)

# Convert the shape of the labels to match the image and select the cluster with max score
labels_mask = labels.reshape(img.shape[:2])
return img * (labels_mask == np.argmax(scores))[:, :, np.newaxis], labels_mask
```

*Code 1: Color segmented image (K = 3)*

After segmentation, the card outlines are detected. This is accomplished by running contour detection on the cards using OpenCV's built in contour detection, relying on Suzuki's Contour Tracing Algorithm[5]. The contours extracted from this image should be the outlines of the cards. This completes the card detection step.

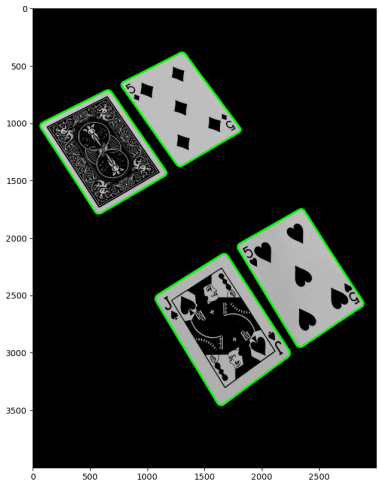

*Figure 3: Contours and mask from segmentation before Ramer-Douglas-Peucker simplification*

After the card outlines are found, they need to be reprojected so they all have the same shape and dimensions. However, since cards have rounded corners, we don't have an obvious set of four points that can be used to solve a homogeneous transformation (see Figure 3). To get these, we attempt to simplify the contours using the Ramer–Douglas–Peucker algorithm[6], also built into OpenCV, which will merge any points closer together than the parameter epsilon. For this process we used an epsilon of 0.02 multiplied by the perimeter of the contour. This epsilon was determined experimentally by looking for a value that doesn't oversimplify or under simplify cards in most cases. The results of the simplification process are shown in Figure 4.

---

[5] Reference [8]

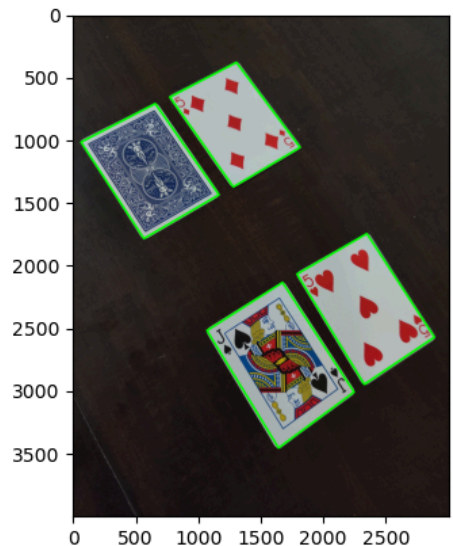

*Figure 4:* *Cards in original image with simplified bounding boxes.*

After simplification, we extract 4 point polygons of sufficiently uniform area, discarding any extreme outliers that are less than 10% the area of the largest polygon. The points are then re-ordered to a known standard order according to an algorithm inspired by the implementation found in PyImageSearch[7]. It sorts by x-coordinate and then extracts the two left-most points. The point with minimum y-coordinate (according to raster coordinates) is assumed to be the top-left point, and the left-most point with maximum y-coordinate is assumed to be the bottom left point. Then, using the fact that the euclidean distance across the diagonal will be the largest as the hypotenuse of a triangle, the two remaining points are sorted according to their distance from the top-left point, with the further one being the bottom right point and the closer one being the top right point. Finally, the points are returned in the order top-left, top-right, bottom-right, and bottom-left. Then the width and height of the card are calculated using the euclidean distances between the points. If the card is landscape, i.e. the width is greater than the height, the card is rotated 90 degrees by rotating the destination points so all cards end up in a portrait orientation. Finally, a homography is then calculated using OpenCV's built in perspective transform functions and applied to the image. This process is repeated for all contours in the image suspected to be cards. The results of this process are shown in Figure 5.

---

[6] Reference [6]
[7] Reference [5]

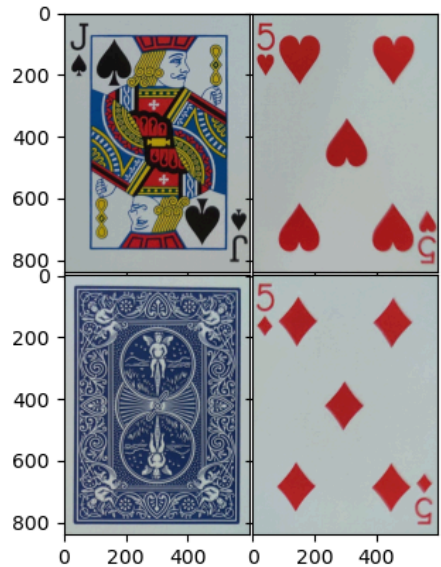

*Figure 5: Reprojected cards*

```python
# https://github.com/PyImageSearch/imutils/blob/9f740a53bcc2ed7eba2558afed8b4c17fd8a1d4c/imutils/perspective.py#L9
def order_points(pts):
        # sort the points based on their x-coordinates
        xSorted = pts[np.argsort(pts[:, 0]), :]
        # grab the leftmost and rightmost points from the sorted x-coordinate points
        leftMost = xSorted[:2, :]
        rightMost = xSorted[2:, :]
        # grab the top-left and bottom-left points, respectively
        leftMost = leftMost[np.argsort(leftMost[:, 1]), :]
        (tl, bl) = leftMost
        # find bottom-right point from max euclidean distance
        D = np.linalg.norm(tl - rightMost, axis=1)
        (br, tr) = rightMost[np.argsort(D)[::-1], :]
        return np.array([tl, tr, br, bl], dtype="float32")

def reproject_playing_card(img):
        # Segmentation (elided)
        # Find contours in the image
        contours, _ = cv.findContours(gray, cv.RETR_EXTERNAL, cv.CHAIN_APPROX_SIMPLE)

        cards = []
        # Loop through the contours and try to turn them into boxes
        for c in contours:
        # Approximate contours to remove extra points
        epsilon = 0.02 * cv.arcLength(c, True)
        contour = cv.approxPolyDP(c, epsilon, True)
        # Probably a playing card if it has 4 corners
        if len(contour) == 4:
            cards.append(contour)

        # Remove outliers (elided)
```

```python
        # Reproject the cards
        norm_cards = []
        for card in cards:
            card = order_points(card.reshape((4, 2)))
            width, height = get_width_height(card)
            # If card is landscape flip the width and height and rotate the dest points
            # Roll numpy array (eliled)

            # Getting the homography
            M = cv.getPerspectiveTransform(
                np.float32(card), np.float32(destination_corners)
            )
            # Perspective transform using homography
            norm_card = cv.warpPerspective(img, M, (width, height), flags=cv.INTER_CUBIC)
            norm_cards.append((card, norm_card))
        return norm_cards
```

***Code 2:*** *Code snippet for card projection (some parts removed for clarity)*

The last step in the pipeline is to classify the card by recognizing the number on the card in the top left corner. To preprocess the input to a standardized format, the corners of each card are cropped, thresholded, and resized to 28x28 pixels. The corner is found using manually tuned values for Bicycle playing cards based on proportions of the card dimensions. This rectangle starts from the row that is 0.03 times the card height and ends at the row that is 0.15 times the card's height and the column 0.01 times the card's width to the column 0.13 times the card's width. In the process, the cards are also verified to have a reasonable aspect ratio for playing cards (around 1.4 ± 0.2), and that the dimensions of the extracted corner are non-zero. The cards are then converted to grayscale and thresholded to extract all pixels with gray values between 0 and 125 using 8-bit color. Lastly, the extracted corner is resized to a standard 28x28 resolution using cubic interpolation. Figure 6 shows the corners of the four cards after the extraction process is complete.

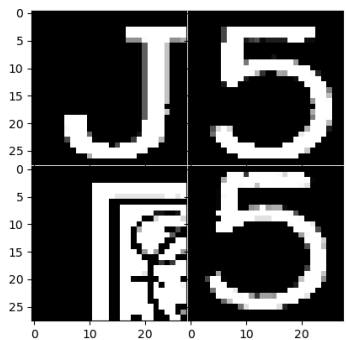

***Figure 6:*** *Corners of all four cards extracted, converted to greyscale, thresholded, and rescaled.*

```python
def extract_corner(card):
```

```python
        h, w, _ = card.shape

        if EXPECTED_CARD_ASPECT - w / h > EPSILON:
            raise RuntimeError(f"invalid card aspect {w / h}, probably invalid")

        # get top left corner of each card
        # just hardcoded, could use contour detection to find biggest thing in a corner
        corner = card[int(0.03 * h) : int(0.15 * h), int(0.01 * w) : int(0.13 * w)]
        ch, cw, _ = corner.shape
        if ch == 0 or cw == 0:
            raise RuntimeError(f"image too small!")

        corner = cv.cvtColor(corner, cv.COLOR_RGB2GRAY)
        corner = cv.inRange(corner, 0, 125)
        corner = cv.resize(corner, (28, 28), interpolation=cv.INTER_CUBIC)
        return corner
```

*Code 3: Code snippet for extracting the corners*

Once the corners have been preprocessed, A K-nearest neighbors (KNN) algorithm is used to recognize the cards. This algorithm compares the corner features to a manually labeled set of training data (see Figure 7) and decides which data type it fits based on its closest neighbors in the feature space. The training images are separate from the ones used to test the algorithm. Instead of doing KNN on the raw pixel data of the corners, all images are fed into a Histogram of Gradients (HoG) descriptor provided by OpenCV. It uses a histogram of the gradients in an image (similar to a SIFT descriptor) to better categorize the features invariant to factors like rotation or scale. This is used to help make the classifier more consistent and resistant to outliers. The descriptor has a window size of 28x28 (the entire image), block size (area over which pixels are normalized) of 14x14, block stride (stride moved between when each block is calculated, in this case overlapping) of 7x7, cell size of 7x7 (the actual cell over which a histogram of gradients is taken), and 9 bins for the histogram. These values are quite arbitrary and many can work well, so long as the same values are used for both the training and classification. The training images and their labels are read from the train directory and the HoG descriptor is applied, after which a KNN is trained on them, using the built-in OpenCV KNN implementation. After the KNN is trained, the new images are fed into it. We used a K value of 3 which means the input image will be compared to its 3 closest neighbors in the feature space. Currently, the KNN is re-trained on every startup, which has minimal performance impact since it only has 176 training examples for all classes. However, as additional training samples are collected, it may need to be saved and loaded as needed.

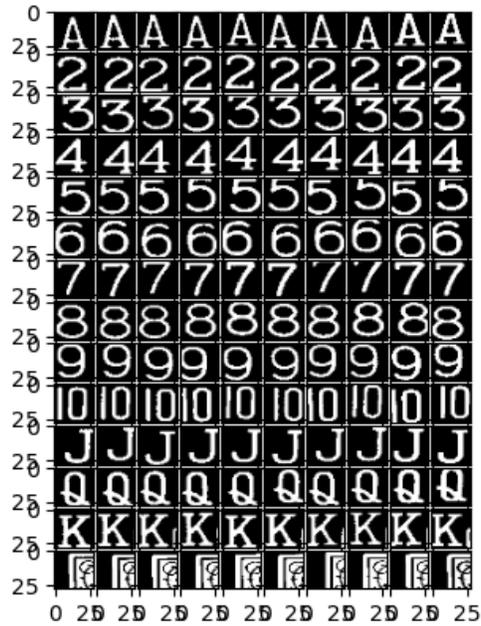

*Figure 7:* *KNN training data. There are 10 different unique examples of each card type that the algorithm can compare to.*

```python
# Initialize HOG descriptor
winSize = (28, 28)
blockSize = (14, 14)
blockStride = (7, 7)
cellSize = (7, 7)
nbins = 9
hog = cv.HOGDescriptor(winSize, blockSize, blockStride, cellSize, nbins)

def extract_hog_features(image):
    # Compute HOG descriptor
    hist = hog.compute(image)
    return hist.flatten()

def train():
    unique_labels = set()
    images = []
    labels = []
    for label_path in TRAIN_DIR.iterdir():
        _, label = label_path.stem.split('-', maxsplit=1)
        for image_path in label_path.iterdir():
            image = cv.imread(str(image_path), cv.IMREAD_GRAYSCALE)
            image = extract_hog_features(image)
            images.append(image)
            labels.append(label)
            unique_labels.add(label)
    unique_labels = sorted(list(unique_labels))
    labels = [unique_labels.index(label) for label in labels]
    knn = cv.ml.KNearest_create()
    knn.train(np.array(images), cv.ml.ROW_SAMPLE, np.array(labels))
    return knn, unique_labels

def predict(knn, labels, image):
    ret, results, neighbours, dist = knn.findNearest(np.array([image]), 3)
```

*Code 4: code snippet for extracting features, training, and performing KNN*

Moving on to the Blackjack Basic Strategy implementation, the algorithm utilizes a rule-based approach to create optimized move recommendations by mapping out and defining the moves for all possible player and dealer card configurations. Since blackjack is a fairly simple game, it has been "solved" meaning that the best move in any scenario is known. Figure 8 shows the full rules table for a single hand of blackjack. The algorithm begins by initializing with player and dealer hands as input parameters, as the program was developed within a separate class structure to make it more versatile and portable. The recommend() method is the core function for determining the optimal move. It first evaluates whether the player has achieved a Blackjack win condition. Next, it considers potential pairs in the player's hand and advises on whether to split or not based on specific card combinations and the dealer's upcard.

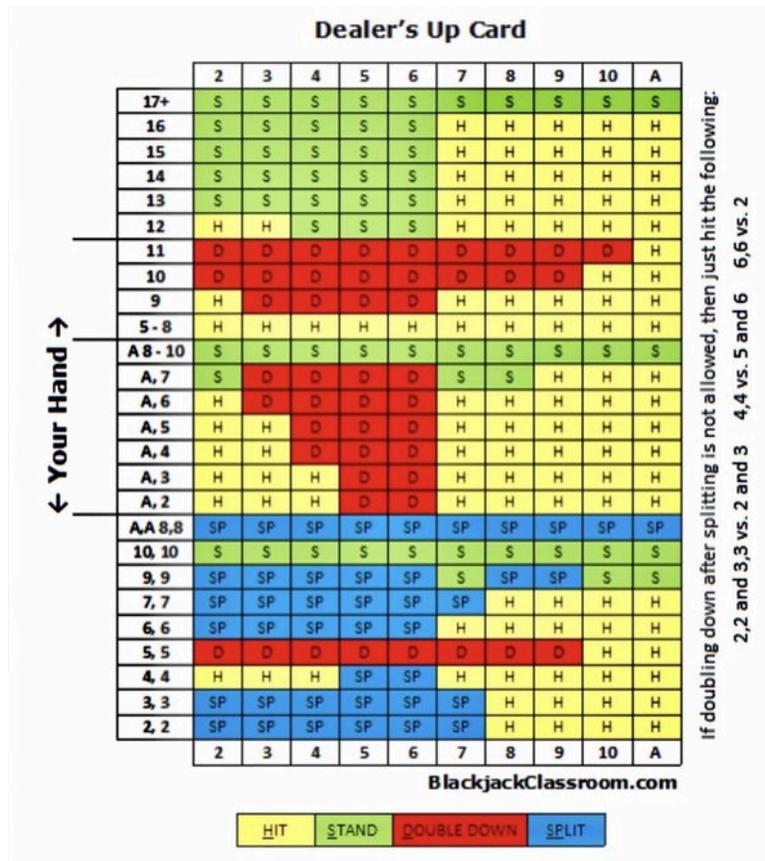

*Figure 8: Known best move for every blackjack hand*

```
def recommend(self):
    # Determine whether to hit, stand, double, or split based on basic strategy.
    # hands will contain 2-10 and A.    J, Q, K already converted to 10
```

```python
    move = ""
    if len(self.player_hand) < 2:
        raise ValueError
    dealer = self.dealer_hand[0]

    # check blackjack win condition
    if self.player_hand == ["A", "10"]:
        return "Blackjack, you win!"

    # check pairs:
    if len(set(self.player_hand)) == 1:
        card = self.player_hand[0]
        move = "Don't split."
        # always split on A/8 pairs
        if card in ("A", "8"):
            move = "Split."
        # never split on 5/10 value pair
        elif card in ("5", "10"):
            move = "Don't split."
        # never split on pairs 2 - 7 if dealer btw 8 - A
        elif 2 <= int(card) <= 7:
            if dealer in ("8", "9", "10", "A"):
                move = "Don't split."
            # split for dealer 2-7 for pairs 2, 3, and 7
            elif card in ("2", "3", "7") and 2 <= int(dealer) <= 7:
                move = "Split."
            # double down split on dealer 2 and 3 for 2 a 3 pairs
            if card in ("2", "3") and dealer in ("2", "3"):
                move = "Double Down Split. If not possible, then hit."
            # For 6 pair, split from dealer 3-6, double down on dealer 2
            if card == "6":
                if dealer.isdigit() and 3 <= int(dealer) <= 6: move = "Split."
                elif dealer == "2": move = "Double Down Split. If not possible, then hit."
            # For 4 pair, double down on dealer 5 and 6. Don't split on dealer 2-4 and 7.
            if card == "4":
                if dealer in ("5", "6"): move = "Double Down Split. If not possible, then hit."
        # last case, 9 pair: split on dealer 2-6, 8, 9. Don't split on 7, 10, A.
        else:
            if dealer not in ("7", "10", "A"): move = "Split."
```

*Code 5: Code snippet for blackjack win condition and pairs logic*

For soft totals, hands containing an Ace, the algorithm considers the value of the Ace and the secondary card to determine the optimal move. It factors in various scenarios, such as when to hit, stand, or double down based on the player's total and the dealer's upcard.

```python
# soft totals (contains an A). We're reverse sorting the list of cards
# so A will be the first element.
elif self.player_hand[0] == 'A':
    second = self.player_hand[1]
    move = "Hit." # default move
    # for soft total 19 or 20, usually stand. Double on soft 19 with dealer 6.
    if second in ("8", "9"):
        if second == "8" and dealer == "6":
            move = "Double."
        else: move = "Stand."
```

```python
        elif second == "7":
            if dealer.isdigit() and 2 <= int(dealer) <= 6:
                move = "Double."
            elif dealer.isdigit() and 7 <= int(dealer) <= 8:
                move = "Stand."
        else:
            if dealer in ("5", "6"):
                move = "Double."
            elif dealer == "4" and second in ("4", "5", "6"):
                move = "Double."
            elif dealer == "3" and second == "6":
                move = "Double."
```

*Code 6: Code snippet for soft totals logic*

For hard totals, hands without an Ace or where the Ace only counts as 1, the algorithm calculates the total value of the hand and recommends appropriate actions based on predefined thresholds and dealer conditions. The calculate_hand_total() method calculates the total value of the player's hand, accounting for the value of Aces and adjusting the total if needed to avoid busting. This avoids the conditions where Ace can be counted as 11.

```python
# hard totals, no A.
    else:
        hard_total = self.calculate_hand_total()
        move = "Hit." # default move
        if hard_total > 17: move = "Stand."
        elif hard_total == 17: move = "Stand."
        elif 13 <= hard_total <= 16 and dealer.isdigit() and 2 <= int(dealer) <= 6:
            move = "Stand."
        elif hard_total == 12 and dealer.isdigit() and 4 <= int (dealer) <= 6: move = "Stand."
        elif hard_total == 11: move = "Double."
        elif hard_total == 10 and dealer.isdigit() and 2 <= int(dealer) <= 9: move = "Double."
        elif hard_total == 9 and dealer.isdigit() and 3 <= int(dealer) <= 6: move = "Double."

    #print(move)
    return move

def calculate_hand_total(self):
    # Calculate the total 'hard' value of a blackjack hand.
    total = 0
    num_aces = 0
    cards = self.player_hand
    cards.sort(reverse=False)
    for card in cards:
        if card.isdigit():
            total += int(card)
        elif card == 'A':
            num_aces += 1
            total += 11

    # Adjust total for aces if needed
    while total > 21 and num_aces > 0:
        total -= 10
```

```
        num_aces -= 1
    return total
```

*Code 7: Code snippet for hard totals logic*

      This complete rules-based system for Blackjack Basic Strategy covers the optimal moves for all the possible player and dealer hand configurations, providing players with the accurate recommendations to maximize their chances of success while playing.

# 5    Analysis

In order to analyze how our blackjack algorithm works, we first need to analyze the results of the CARP algorithm. Since the algorithm was designed to be somewhat flexible, we decided to test it on many different types of images of cards, not just blackjack hands. Figure 9 shows the output from the program on an image of many playing cards. We find a 100% success rate for the cards that are fully within frame! This demonstrates that the algorithm is not constrained to 4 cards. This also shows that our algorithm is scale invariant because the cards in this image have a much smaller pixel area than the cards used in the training set. The reprojection step ensures that all detected cards will be normalized to the same size before further analysis. Finally, this shows that the algorithm can accurately identify any card regardless of suit.

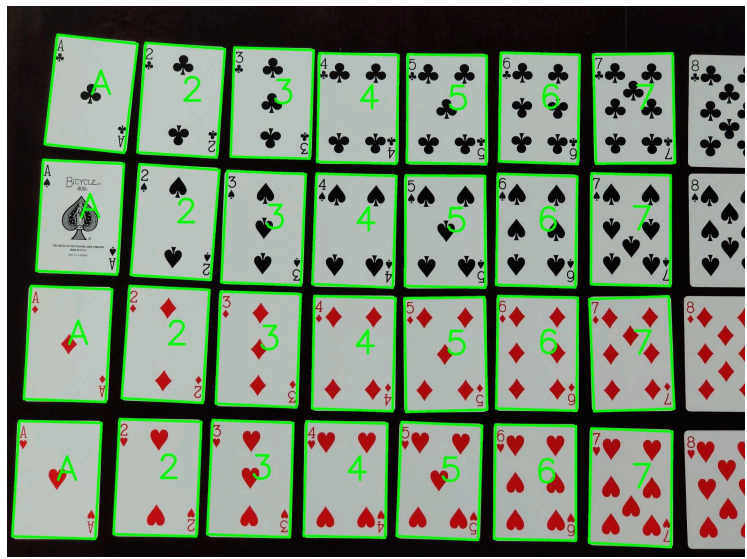

*Figure 9:* Output of CARP from an input similar to the training data.

This image does show one limitation of our approach. It often fails to recognize the cards that are cut off on the side of the image or are otherwise not quadrilaterals. Figure 10 shows another result from a more difficult image. This image was chosen to demonstrate the rotation invariance as well as the ability of the reprojection algorithm to handle skewed cards. This image has a 95% accuracy rate for the cards that it detected. The only card it misclassified was the 9 of spades which was cut off by the top of the image so we don't expect our algorithm to be able to identify it anyways. This image also shows that CARP cannot handle cards that are overlapping. The clusters for these instances have a non-quadrilateral shape so they are rejected by the algorithm. This is probably the biggest downside of our approach because it is common for players to keep their cards overlapping. However, for our purpose this isn't a big problem because this algorithm is designed for educational purposes.

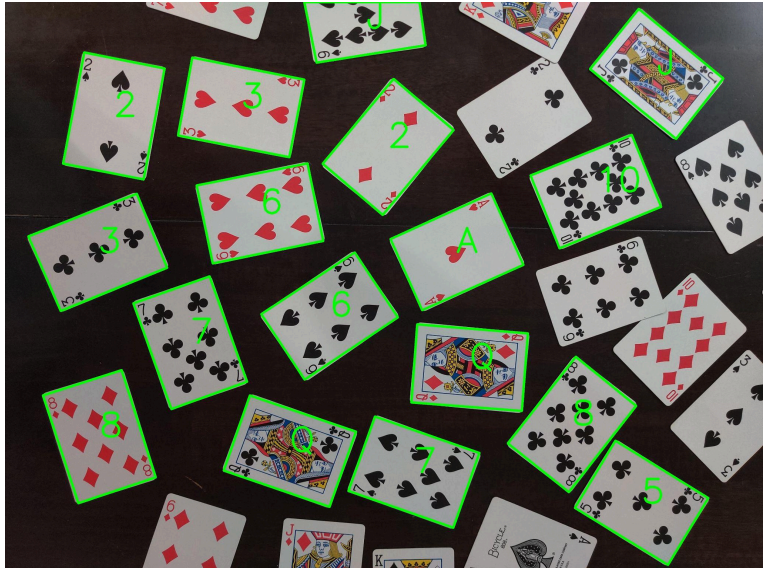

*Figure 10:* Output of CARP from an image that is significantly different from the training data

Finally, we tested on a skewed and rotated image of a more realistic blackjack setup. Figure 11 shows how effective our algorithm is at analyzing ideal blackjack setups. We ended up testing on about a dozen images of blackjack setups and we achieved a 100% success rate on these types of images. When testing on a large data set of images like Figures 9 and 10 we found a very high accuracy of 91%. The confusion matrix in Figure 12 shows the total accuracy of each type of card and all misclassifications.

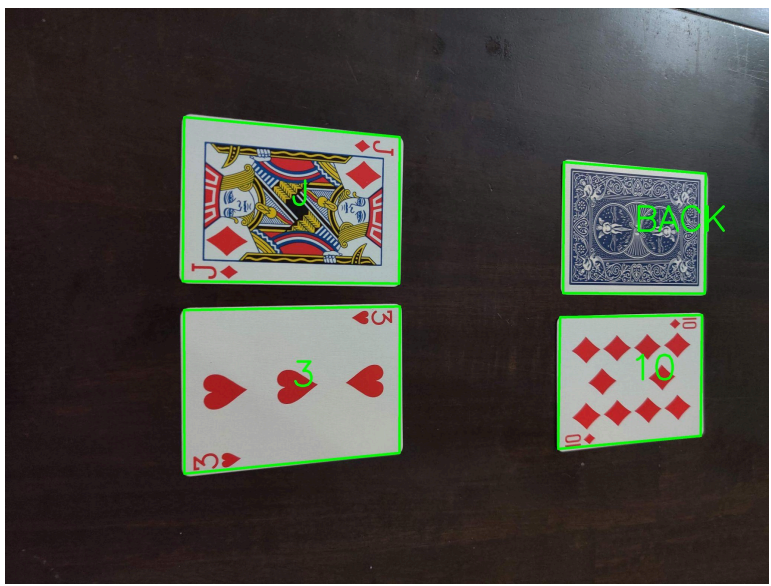

*Figure 11:* Output of CARP for a skewed image of a blackjack setup

In order to more precisely quantify the accuracy of the classification algorithm, we gathered the total precision, recall, f1-score, and support for each type of card. Table 1 shows this data. The precision score is the number of cards correctly classified divided by the total number of cards classified as that type. So a precision score of 1 means that there were no false positives. The recall score is the number of correctly classified cards divided by the actual number of cards of that type. This is a measure of how many false negatives there were. The f1-score is a combination of these two so a high f1-score score means the algorithm performed well. Finally, the support score shows how many cards of this type were tested. At the bottom of the table, the total accuracy score is a weighted average of the f1-scores. The macro average is the unweighted averages of all scores in the column. The total support is the total number of cards tested. Overall, an accuracy over 90% is quite good. To improve this, more training and testing data would be needed.

| Card | precision | recall | f1-score | support |
|---|---|---|---|---|
| 10 | 1.00 | 1.00 | 1.00 | 10 |
| 2 | 1.00 | 0.90 | 0.95 | 10 |
| 3 | 0.88 | 1.00 | 0.93 | 7 |
| 4 | 1.00 | 1.00 | 1.00 | 5 |
| 5 | 0.91 | 0.91 | 0.91 | 11 |
| 6 | 0.92 | 1.00 | 0.96 | 12 |
| 7 | 1.00 | 0.89 | 0.94 | 9 |
| 8 | 1.00 | 1.00 | 1.00 | 10 |
| 9 | 0.83 | 1.00 | 0.91 | 5 |
| A | 1.00 | 1.00 | 1.00 | 10 |
| BACK | 0.92 | 0.60 | 0.73 | 20 |
| J | 0.56 | 1.00 | 0.72 | 9 |
| K | 1.00 | 0.75 | 0.86 | 4 |
| Q | 1.00 | 1.00 | 1.00 | 9 |
| Accuracy | | | 0.91 | 131 |
| macro avg | 0.93 | 0.93 | 0.92 | 131 |
| weighted avg | 0.93 | 0.91 | 0.91 | 131 |

*Table 1:* KNN evaluation data

     Figure 12 shows the confusion matrix for the KNN algorithm. The algorithm does a really good job at identifying most of the cards, there are a few seemingly random misclassifications, but the most interesting thing is that it frequently misclassifies a card back as a jack but not the other way around. This may be because the card backs have strong vertical and horizontal elements and all of their pixels are on the right side of the window similar to the J. Most testing is needed to determine the exact cause of this failure. It would also be nice to gather more test data for 4s, 9s, and Ks. Since the images all had random cards, the frequency of these cards happened to be lower than the others.

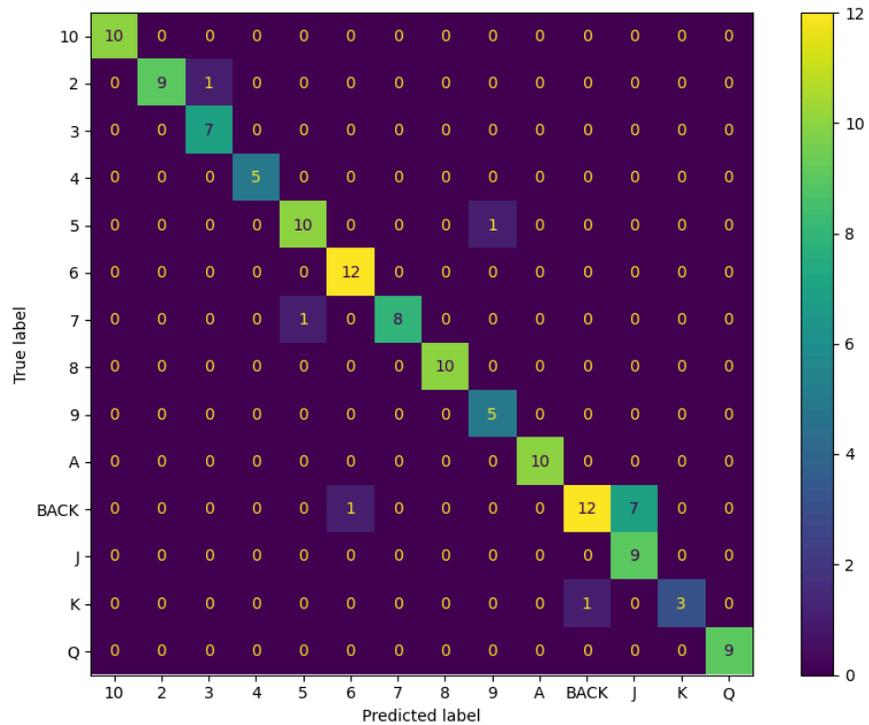

*Figure 12: Confusion Matrix for KNN*

The blackjack algorithm was very easy to test. The main question we needed to answer was: does it predict the right moves? After testing for all of the standard blackjack images, we found that it predicted the right move 100% of the time. Figure 13 shows example outputs for a bunch of different scenarios including different best moves and different camera angles. This shows that the full algorithm is very robust and great at doing what it is designed to do.

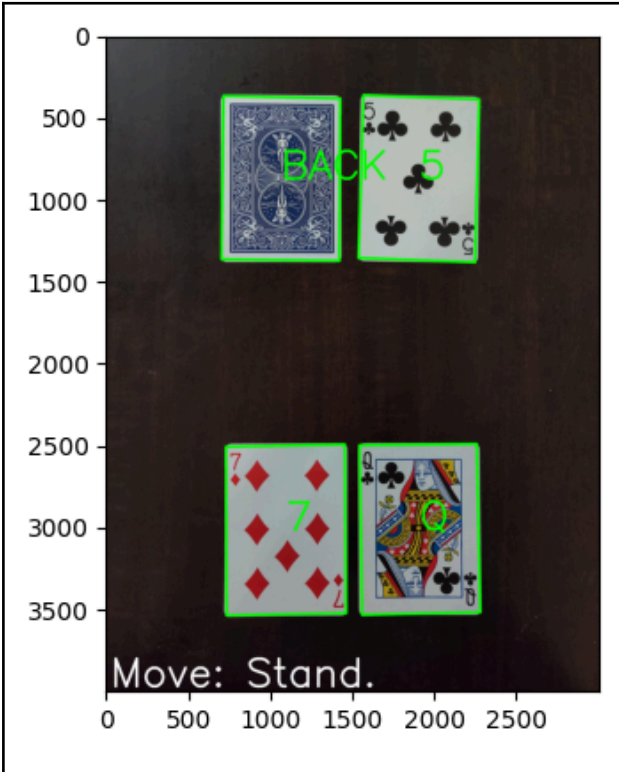
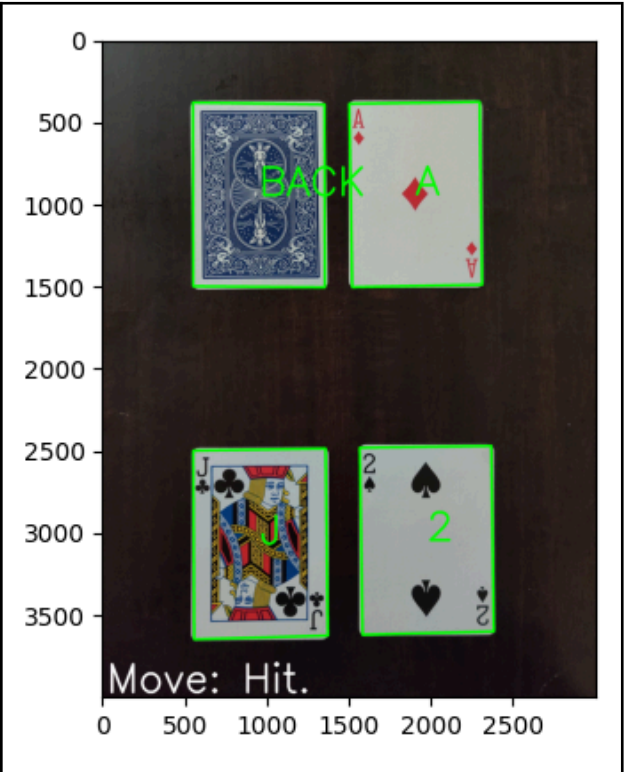
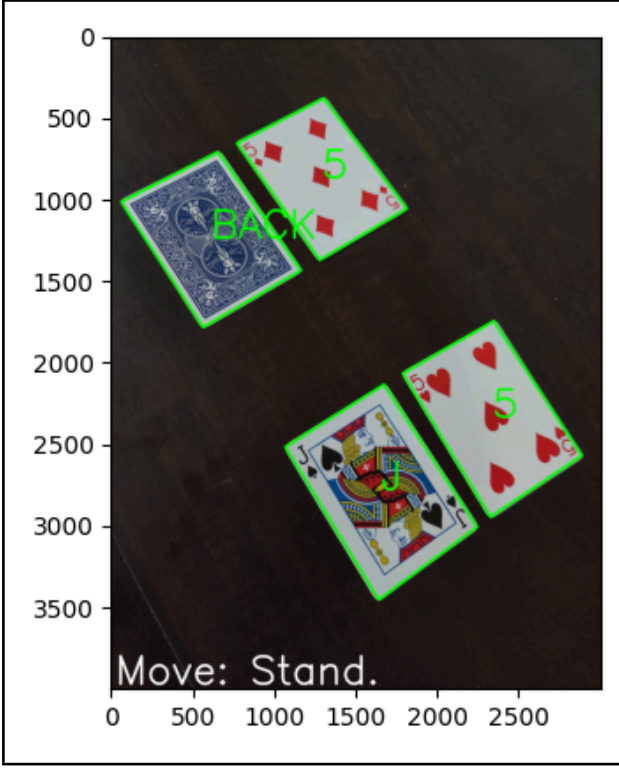
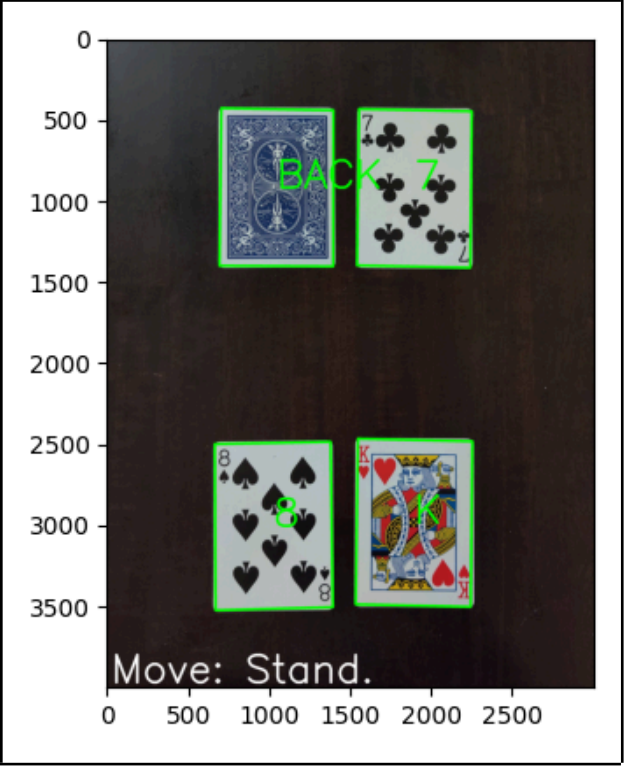

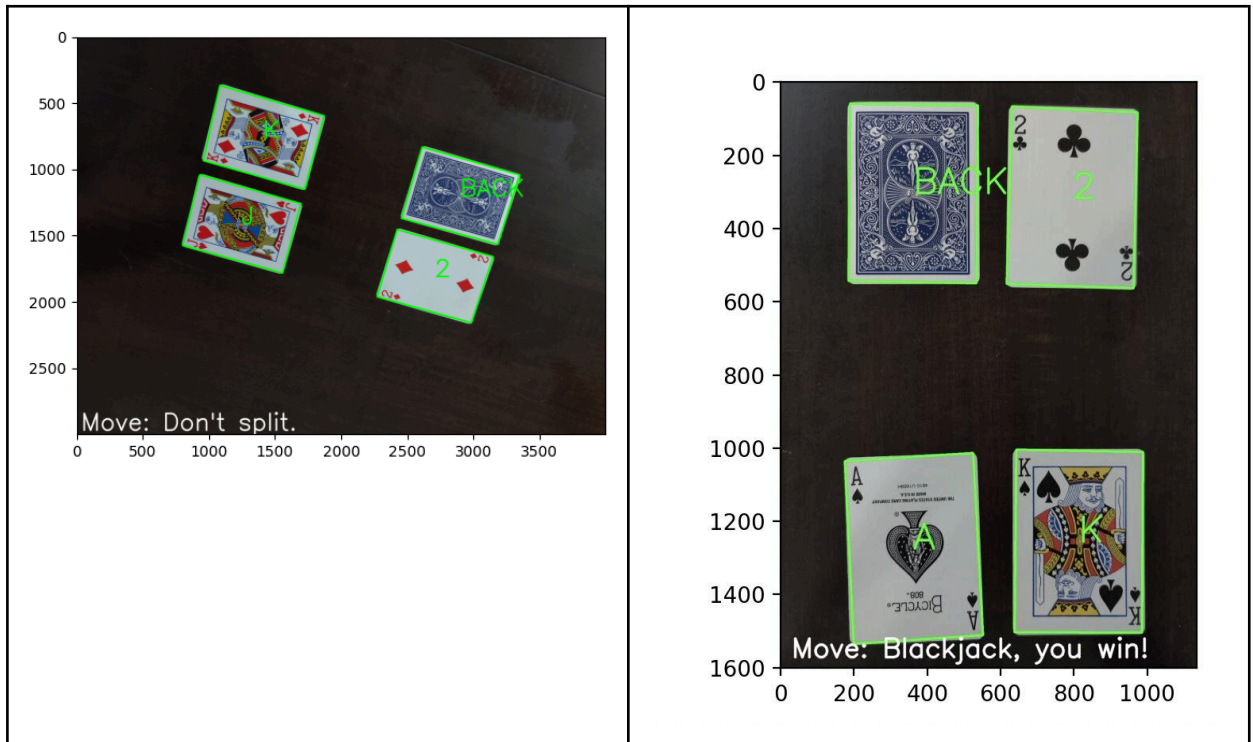

*Figure 13: Output of the full program including blackjack move recommendation.*

Since the blackjack algorithm only uses information about the current hand, this technique will still lose to the casino on average. In order to improve our odds, we would need to introduce card counting. Card counting is a technique of keeping track of all the previous cards that have been played and continuously updating the probabilities of the next card to improve the betting strategy. This would be much more difficult to implement and would require the program to be able to run multiple times in a row and store all previous hands. There are also levels to card counting. Strategies used by human players involve updating a tally based on if a high or low card has been played. However, this doesn't use all the information available. In theory, our program could remember the exact cards that have been played and play perfectly.

# Summary

The card classification algorithm uses K-means with K=3 to separate the cards from the background because this worked the best out of the techniques we tried. Then it uses a reprojection algorithm to get the cards into a standardized size and orientation. This allows us to classify a wide variety of images. Then it crops in on the corner containing the number and thresholds it to obtain a simple standardized format. Finally it uses KNN to classify the cards based on a large set of training data. The blackjack algorithm separates the player's cards from the dealer's cards and uses a known best strategy for single hands of blackjack.

For being completely classically based, this algorithm for card detection and blackjack move prediction works amazingly well achieving an overall accuracy of 100% for blackjack images tested and 91% for other images of cards. No other paper has managed to achieve such good results for card detection using purely classical techniques. This is in part due to the extensive preprocessing done to the images to achieve consistent card numbers for feature extraction and classification. However, this algorithm only works under the specific conditions it was designed for. It cannot handle overlapping cards and the images have to be lit well and uniformly enough for the K-means algorithm to produce good results. This program could be further improved with more training data and more testing.

Since the blackjack move prediction is based on a known perfect blackjack move table, it will always produce the best results for a single hand. However, a really good blackjack player might still be able to beat this algorithm if they are good at counting cards because this algorithm assumes all cards are equally likely. A better version of this algorithm would allow the user to input multiple images in a row, count the cards, and adjust the moves based on the probabilities of the next possible cards.